\documentclass[sigconf]{acmart}
\AtBeginDocument{%
  \providecommand\BibTeX{{%
    \normalfont B\kern-0.5em{\scshape i\kern-0.25em b}\kern-0.8em\TeX}}}

\setcopyright{rightsretained}
\copyrightyear{2024}
\acmYear{2024}
\acmDOI{XXXX.XXXX}

\acmConference[Designing an Intro to HRI Course Workshop at HRI 2024]{Presented at the Designing an Intro to HRI Course Workshop at HRI 2024 (arXiv:2403.05588) Report-no: HRI101/2024/10}{March 11, 2024}{Boulder, CO}
\begin{document}

\title[UChicago Course ``Human-Robot Interaction: Research and Practice'']{Teaching Introductory HRI: UChicago Course\\ ``Human-Robot Interaction: Research and Practice''}

\author{Sarah Sebo}
\email{sarahsebo@uchicago.edu}
\orcid{0000-0003-2211-6429}
\affiliation{%
  \institution{The University of Chicago}
  \city{Chicago}
  \country{USA}
}

\renewcommand{\shortauthors}{Sebo}

\begin{abstract}
  In 2020, I designed the course CMSC 20630/30630 Human-Robot Interaction: Research and Practice as a hands-on introduction to human-robot interaction (HRI) research for both undergraduate and graduate students at the University of Chicago. Since 2020, I have taught and refined this course each academic year. Human-Robot Interaction: Research and Practice focuses on the core concepts and cutting-edge research in the field of human-robot interaction (HRI), covering topics that include: nonverbal robot behavior, verbal robot behavior, social dynamics, norms \& ethics, collaboration \& learning, group interactions, applications, and future challenges of HRI. Course meetings involve students in the class leading discussions about cutting-edge peer-reviewed research HRI publications. Students also participate in a quarter-long collaborative research project, where they pursue an HRI research question that often involves conducing their own human-subjects research study where they recruit human subjects to interact with a robot. In this paper, I detail the structure of the course and its learning goals as well as my reflections and student feedback on the course. 
\end{abstract}





\maketitle

\section{Introduction}

At the University of Chicago, I created the course CMSC 20630/30630 Human-Robot Interaction: Research and Practice to serve as an introductory course for both undergraduate and graduate students interested in learning more about and gaining hands-on experience in human-robot interaction (HRI). Students are exposed to important topics in HRI through a carefully selected reading list of HRI publications, where most class periods are devoted to a class discussion of one of these papers. Students also gain hands-on experience with HRI by conducting an HRI research project of their own design that they complete throughout the academic quarter. 

From my observations and the feedback I have received from students, this course has been well-received. I, as the instructor, am able to broaden and deepen my thoughts about HRI through thoughtful class discussions and gain new inspiration for research ideas and directions through students' course projects. Students reported enjoying the class discussions, feeling challenged to think about human-robot interaction in new ways. They also gained tremendously from the course project, which gave them real, tangible experience conducting HRI research. 

This paper focuses on describing the structure of the course and my rationale for its design. I also include some student feedback about the course and my own reflections on the strengths of the course and areas for improvement. For further details, please refer to the \textbf{\href{https://classes.cs.uchicago.edu/archive/2024/winter/20630-1/index.html}{2024 course website}\footnote{https://classes.cs.uchicago.edu/archive/2024/winter/20630-1/index.html}} and my \textbf{\href{https://sarahsebo.com/teaching.html}{personal website}\footnote{https://sarahsebo.com/teaching.html}} for links to all of my publicly available course materials. 

\section{Course Structure}

At the University of Chicago, CMSC 20630/30630 Human-Robot Interaction: Research and Practice enrolls \textbf{a maximum of 20 students} in order to enable an environment for fruitful in-class discussions where all students are able to contribute. While this class size may not scale to other institutions, this is a typical class size for many specialized Computer Science courses at the University of Chicago. Additionally, the University of Chicago operates on the quarter system, where each quarter involves 9 weeks of instruction  (semester schedules typically include ~13 weeks of instruction) and 1 week for finals. In the rest of this section I detail the structure of the course, its learning objectives, the format of class meetings, and details about the quarter-long class project. 

\subsection{Course Description}

Robots are increasingly common in our everyday spaces: tutoring elementary students, assisting human workers in manufacturing contexts, providing museum tours, interacting with families within their homes, and helping to care for the elderly. One critical factor to the success of these robots is their ability to effectively interact with people: human-robot interactions.

This course focuses on the core concepts and cutting-edge research in the field of human-robot interaction (HRI), covering topics that include: nonverbal robot behavior, verbal robot behavior, social dynamics, norms \& ethics, collaboration \& learning, group interactions, applications, and future challenges of HRI. Course meetings involve students in the class leading discussions about cutting-edge peer-reviewed research HRI publications. Students also participate in a quarter-long collaborative research project, where they pursue an HRI research question that often involves conducing their own human-subjects research study where they recruit human subjects to interact with a robot.

The course has no prerequisites, however backgrounds in robotics or HCI is encouraged. Programming knowledge is not necessary, however, is often useful for the course project.

\subsection{Learning Objectives}

This course has three central learning objectives:
\begin{enumerate}
    \item Students will obtain a broad understanding and exposure to cutting-edge research in the field of Human-Robot Interaction through course readings and discussions of those readings.
    \item Students will cultivate analytical and critical thinking skills when evaluating research in HRI through leading class discussions and making analytical comments on the readings.
    \item Students will gain hands-on HRI research experience through the course project.
\end{enumerate}

\subsection{Class Meetings: Discussions about HRI Research Papers}

Class meetings occur three times a week (typically Monday, Wednesday, Friday) for 50 minutes each. As shown in Table~\ref{tab:paper-schedule}, each week of the course is focused on a different core topic in HRI. During each class meeting, the class reads and discusses one paper. I carefully select these papers to include a distribution of:
\begin{itemize}
    \item Foundational papers in the field (e.g., \citet{robinette2016overtrust}'s investigation of people's overtrust of robots in emergency evacuation scenarios, \citet{brvsvcic2015escaping}'s exploration of robot abuse in shopping malls, \citet{kahn2012robovie}'s examination of child responses to a robot being forced to go in a closet)
    \item HRI papers published within the last year (e.g., for the Winter 2024 version of the course I included 8 papers published in 2023, 6 of which were published at the HRI 2023 conference~\cite{cui2023no, brawer2023interactive, jeong2023robotic, ligthart2023design, nanavati2023design, winkle2023feminist} and 2 of which were published in other HRI or robotics publication venues~\cite{mavrogiannis2022winding, natarajan2023human})
    \item Papers that cover important topics in the field of HRI (e.g., trust~\cite{robinette2016overtrust}, nonverbal robot behaviors such as gaze~\cite{gillet2021robot} and gestures~\cite{rifinski2021human}, HRI ethics~\cite{winkle2023feminist}, multiparty HRI~\cite{erel2022carryover, reig2021flailing, tennent2019micbot})
\end{itemize}

Before each class meeting, each student is expected to read the assigned paper and submit at least one \textbf{analytical comment} on the paper using the Google Chrome extension \href{https://web.hypothes.is/}{hypothes.is}. Analytical comments are intended to demonstrate critical thinking about the paper and can, for example, highlight and justify strengths or weaknesses with the methodology or data analysis, suggest a fruitful avenue for future research, discuss potential implications. 

In addition to the required analytical comment, 2-3 students are asked to write a \textbf{peer-review} of the paper.
Guidelines for the peer-review are similar to those used for the ACM/IEEE International Conference on Human-Robot Interaction. Each student is asked to write two peer reviews during the course, which are due before the class discussion of that paper. 

During each class meeting one student serves as the \textbf{discussion leader}. The discussion leader gives a brief presentation, where they provide an overview of the paper, highlight strengths and weaknesses, and pose discussion questions for the class. Then, the discussion leader moderates a class discussion. 

For 2-3 class periods during the quarter, I invite the first author of one of the recently published papers to join our class as a \textbf{guest presenter}. The guest presenter (via Zoom) presents their paper and then engages in an open question and answer session with the students. This provides students with the opportunity to directly learn from top researchers in the field and gain a deeper insight into the work that goes into the papers we read in class. 

\begin{table}
  \caption{Class Meeting Schedule for UChicago's HRI: Research \& Practice in the Winter Quarter of 2024}
  \label{tab:paper-schedule}
  \begin{tabular}{lll}
    \toprule
    Week&Topic&Paper Sub-Topics \& Citations\\
    \midrule
    1 & Course Introduction & Robot Embodiment~\cite{bainbridge2011benefits}\\
    2 & Verbal Behavior & Emotion~\cite{pelikan2020you}\\
    & & Self-Disclosure~\cite{martelaro2016tell}\\
    3 & Nonverbal Behavior & Gaze~\cite{gillet2021robot}\\
    & & Non-humanoid Gestures~\cite{rifinski2021human}\\
    4 & Social Dynamics & Trust~\cite{robinette2016overtrust}\\
    & & Conflict Resolution~\cite{babel2022will}\\
    & & Robot Role~\cite{jeong2023robotic}\\
    5 & Norms/Ethics & Harsh Robot Treatment~\cite{kahn2012robovie}\\
    & & Social Norms~\cite{komatsu2021blaming}\\
    & & Ethical Research Practice~\cite{winkle2023feminist}\\
    6 & Collaboration \& & Crowd Navigation~\cite{mavrogiannis2022winding}\\
    &  Learning & Language Corrections~\cite{cui2023no}\\
    & & Interactive Policy Shaping~\cite{brawer2023interactive}\\
    7 & Group Interactions & Shaping Group Dynamics~\cite{tennent2019micbot} \\
    & & Carryover Effects~\cite{erel2022carryover}\\
    & & Multiple Robots~\cite{reig2021flailing}\\
    8 & Applications & Public Spaces~\cite{brvsvcic2015escaping}\\
    & & Education~\cite{ligthart2023design}\\
    & & Robot-Assisted Feeding~\cite{nanavati2023design}\\
    9 & Grand Challenges & Human-Robot Teaming~\cite{natarajan2023human}\\
  \bottomrule
\end{tabular}
\end{table}

\subsection{Hands-On Course Project where Students Conduct their own HRI Research Project}

The hands-on course project is the cornerstone of this course, where students conduct a quarter-long HRI research project of their own design. Students complete their projects in groups of 2-4 and are given access to my research lab's resources (i.e., robots) to complete their projects. The robots available to students for their course projects in my lab include: 2 SoftBank Robotics Nao robots, 1 Hello Robot Stretch robot, 1 Franka Emika Panda robot, 7 Anki Vector Robots, 2 Misty Robotics Misty II robots, 11 ClicBot robots, and 12 Turtlebot3 robots. 

Students are told that their project must (1) be focused on the interaction between physically embodied robots and one or more people and (2) have the potential to contribute new knowledge to the field of HRI. While most students choose to conduct an in-person human subjects study, students are allowed to conduct a project that falls under one of the following categories: 
\begin{itemize}
    \item \textbf{An in-person human-subjects study}: this option gives students the chance to conduct an in-person human-subjects study of their own design. Students are given target number of participants to recruit based on their group size (e.g., 10 participants for groups of 2, 24 participants for groups of 4). 
    \item \textbf{A human-subjects study evaluated on an online crowd-sourced platform}: this option gives students the chance to conduct an HRI human-subjects study on an online crowd-sourced platform (e.g., Prolific), that would enable them to gather a larger sample size. 
    \item \textbf{Analysis of a human-robot interaction in a pre-recorded dataset}: this option requires students to find an HRI dataset and analyze the data in a novel way.
    \item \textbf{Development of a computational model to improve specific human-robot interactions}: this option involves the development of a novel computational model, algorithm, or framework to improve human-robot interactions. Students pursuing this option are expected to both build a novel computational tool for HRI and evaluate it against current state-of-the-art approaches. 
    \item \textbf{A literature review}: this option involves diving into the published work on a particular topic, conducting a systematic literature review, analyzing the papers, and writing a review that both presents the work in the chosen area and highlighting opportunities for future work. 
\end{itemize}

\begin{table}
  \caption{Typical course project schedule for teams pursuing an in-person human-subjects study}
  \label{tab:project-schedule}
  \begin{tabular}{lp{7cm}}
    \toprule
    Week&Course Project Activities\\
    \midrule
    1 & Introduction of course project\\
    2 & Students develop 1-3 course project ideas, pitch them to the class, and form project teams\\
    3 & Initial project proposal is due, the instructor meets with each team to provide feedback on the projects\\
    4 & Finalized project proposal is due that incorporates instructor feedback, students finalize their study design\\
    5 & Students submit an Institutional Review Board (IRB) application for their study\\
    6 & Progress report \#1 is due, students work to program the robot for their study\\
    7 & Students begin to pilot their studies with the instructor and others\\
    8 & Progress report \#1 is due, students recruit and run human subjects\\
    9 & Students finish running human subjects, analyze their data, and make a final presentation to the class\\
    10 & Students finalize their projects and submit a final report\\
  \bottomrule
\end{tabular}
\end{table}

Students are expected to complete an end-to-end HRI project. For those who choose to pursue a human-subjects study, this entails: developing a study design, devloping appropriate measures, submitting an Institutional Review Board (IRB) application, programming a robot, recruiting and running 10-24 human subjects, analyzing the resulting data with appropriate statistical tests, and writing a final report that resembles an HRI conference paper. An example project timeline is shown in Table~\ref{tab:project-schedule}.

The teaching staff (instructor, TA) provide students with considerable support on their course projects. We meet with students (often several times) to refine study designs, review IRB applications, provide students with help getting set up and debugging robots, pilot students' studies before they run participants, help distribute human subjects payments, and provide guidance and example R code for how to conduct statistical analyses of HRI data. 

\subsection{Grading}

This course is graded as follows: 
\begin{itemize}
    \item 60\% project
    \item 20\% analytical comments on the readings
    \item 10\% discussion leadership \& peer-reviews of readings
    \item 10\% attendance/participation
\end{itemize}

Students are provided with a final project grading rubric, so they are aware of how the project grade is allocated. Of the students grade that comes from the project, 50\% comes from project work, 25\% from the final report, 15\% from intermediary deliverables, and 10\% from their final project presentation. 




\section{Student Feedback}

This course has been well received by students, with 100\% of Human-Robot Interaction: Research and Practice students agreeing with ``I would recommend this course to other students'' in UChicago's student course evaluations. Below, I report some common themes in the course feedback. 





\subsection{Vibrant Class Discussions}

Students thought that the paper readings and class discussions were an effective way to learn about HRI, with students saying ``\textit{asking students to read papers and lead presentations is a good way to lean the key methods and know about HRI research}'' and ``\textit{the course was great and the readings were a good summary of HRI.}'' 
While class meeting consisting entirely of discussions may have initially sounded dull to students, students reported having ``\textit{genuinely interesting}'' and ``\textit{thought provoking}'' discussions. Students found class meetings ``\textit{went by quickly and often led to post–class discussions that would linger out into the hallway}'' and that ``\textit{it was quite rare that we would run out of things to talk about when discussing a paper.}''







\subsection{Appreciation of Hands-On Final Project}

Students appreciated the chance to do a hands-on project in the course, gaining valuable experience both in HRI and in conducing research. Students described the course project as ``\textit{the highlight of the course}'' and that the ``\textit{practical research experience [was] very valuable}.'' One undergraduate student's experience in particular highlights the value of the hands-on research project, who said, ``\textit{I genuinely appreciated this project a lot. As an economics major my prior experience in the field of HRI could generously be described as limited. So really getting the opportunity to get my feet wet with a research experience through the project was very exciting for me}.''







\subsection{Desire for More Instruction on HRI Methods}

Some students noted that it could have been helpful for more class time to have been spent the methods of conducting HRI research. For example, one student said that ``\textit{having one or two class sessions on concrete skills related to the class project, such as conducting data analysis or designing surveys, would be helpful}.''



\section{Reflections on Strengths and Areas of Improvement for the Course}

I believe that the core strength of this course is the hands-on HRI research project that students conduct through the course of the quarter. Students have the opportunity to truly see how people interact with robots, can test their own hypotheses about how they believe these human-robot interactions will go, and gain experience with every aspect of the HRI research process. In fact, several of these projects have later continued on and have either been published~\cite{lin2022benefits, mazursky2022physical, naggita2022parental} or are in the submission process. The in-class discussions are also a strength of the course, and provide a format for ideas to be exchanged in an engaging and respectful format. 

With respect to areas for improvement, students do report the desire for additional support and instruction on HRI methods. In the four years I have taught this class, I have approached it with the idea that these skills are better learned ``on the job'' when you can directly apply them to the course project. However, there may be opportunities for these methods to be more formally taught in a way that can help students easily apply these skills to their course projects without such a high learning barrier. 


\begin{acks}
I thank the students who have taken this course, who have engaged meaningfully in class discussions, changed how I think about about HRI, and helped to improve of this course each year. I also thank Brian Scassellati, whose CPSC 473 Intelligent Robotics Lab course has greatly influenced the design of this course. 
\end{acks}

\bibliographystyle{ACM-Reference-Format}
\bibliography{hri_course}


\begin{thebibliography}{24}


\ifx \showCODEN    \undefined \def \showCODEN     #1{\unskip}     \fi
\ifx \showDOI      \undefined \def \showDOI       #1{#1}\fi
\ifx \showISBNx    \undefined \def \showISBNx     #1{\unskip}     \fi
\ifx \showISBNxiii \undefined \def \showISBNxiii  #1{\unskip}     \fi
\ifx \showISSN     \undefined \def \showISSN      #1{\unskip}     \fi
\ifx \showLCCN     \undefined \def \showLCCN      #1{\unskip}     \fi
\ifx \shownote     \undefined \def \shownote      #1{#1}          \fi
\ifx \showarticletitle \undefined \def \showarticletitle #1{#1}   \fi
\ifx \showURL      \undefined \def \showURL       {\relax}        \fi
\providecommand\bibfield[2]{#2}
\providecommand\bibinfo[2]{#2}
\providecommand\natexlab[1]{#1}
\providecommand\showeprint[2][]{arXiv:#2}

\bibitem[Babel et~al\mbox{.}(2022)]%
        {babel2022will}
\bibfield{author}{\bibinfo{person}{Franziska Babel}, \bibinfo{person}{Philipp Hock}, \bibinfo{person}{Johannes Kraus}, {and} \bibinfo{person}{Martin Baumann}.} \bibinfo{year}{2022}\natexlab{}.
\newblock \showarticletitle{It Will not take long! Longitudinal effects of robot conflict resolution strategies on compliance, acceptance and trust}. In \bibinfo{booktitle}{\emph{2022 17th ACM/IEEE International Conference on Human-Robot Interaction (HRI)}}. IEEE, \bibinfo{pages}{225--235}.
\newblock


\bibitem[Bainbridge et~al\mbox{.}(2011)]%
        {bainbridge2011benefits}
\bibfield{author}{\bibinfo{person}{Wilma~A Bainbridge}, \bibinfo{person}{Justin~W Hart}, \bibinfo{person}{Elizabeth~S Kim}, {and} \bibinfo{person}{Brian Scassellati}.} \bibinfo{year}{2011}\natexlab{}.
\newblock \showarticletitle{The benefits of interactions with physically present robots over video-displayed agents}.
\newblock \bibinfo{journal}{\emph{International Journal of Social Robotics}}  \bibinfo{volume}{3} (\bibinfo{year}{2011}), \bibinfo{pages}{41--52}.
\newblock


\bibitem[Brawer et~al\mbox{.}(2023)]%
        {brawer2023interactive}
\bibfield{author}{\bibinfo{person}{Jake Brawer}, \bibinfo{person}{Debasmita Ghose}, \bibinfo{person}{Kate Candon}, \bibinfo{person}{Meiying Qin}, \bibinfo{person}{Alessandro Roncone}, \bibinfo{person}{Marynel V{\'a}zquez}, {and} \bibinfo{person}{Brian Scassellati}.} \bibinfo{year}{2023}\natexlab{}.
\newblock \showarticletitle{Interactive Policy Shaping for Human-Robot Collaboration with Transparent Matrix Overlays}. In \bibinfo{booktitle}{\emph{Proceedings of the 2023 ACM/IEEE International Conference on Human-Robot Interaction}}. \bibinfo{pages}{525--533}.
\newblock


\bibitem[Br{\v{s}}{\v{c}}i{\'c} et~al\mbox{.}(2015)]%
        {brvsvcic2015escaping}
\bibfield{author}{\bibinfo{person}{Dra{\v{z}}en Br{\v{s}}{\v{c}}i{\'c}}, \bibinfo{person}{Hiroyuki Kidokoro}, \bibinfo{person}{Yoshitaka Suehiro}, {and} \bibinfo{person}{Takayuki Kanda}.} \bibinfo{year}{2015}\natexlab{}.
\newblock \showarticletitle{Escaping from children's abuse of social robots}. In \bibinfo{booktitle}{\emph{Proceedings of the tenth annual acm/ieee international conference on human-robot interaction}}. \bibinfo{pages}{59--66}.
\newblock


\bibitem[Cui et~al\mbox{.}(2023)]%
        {cui2023no}
\bibfield{author}{\bibinfo{person}{Yuchen Cui}, \bibinfo{person}{Siddharth Karamcheti}, \bibinfo{person}{Raj Palleti}, \bibinfo{person}{Nidhya Shivakumar}, \bibinfo{person}{Percy Liang}, {and} \bibinfo{person}{Dorsa Sadigh}.} \bibinfo{year}{2023}\natexlab{}.
\newblock \showarticletitle{No, to the Right: Online Language Corrections for Robotic Manipulation via Shared Autonomy}. In \bibinfo{booktitle}{\emph{Proceedings of the 2023 ACM/IEEE International Conference on Human-Robot Interaction}}. \bibinfo{pages}{93--101}.
\newblock


\bibitem[Erel et~al\mbox{.}(2022)]%
        {erel2022carryover}
\bibfield{author}{\bibinfo{person}{Hadas Erel}, \bibinfo{person}{Elior Carsenti}, {and} \bibinfo{person}{Oren Zuckerman}.} \bibinfo{year}{2022}\natexlab{}.
\newblock \showarticletitle{A carryover effect in hri: Beyond direct social effects in human-robot interaction}. In \bibinfo{booktitle}{\emph{2022 17th ACM/IEEE International Conference on Human-Robot Interaction (HRI)}}. IEEE, \bibinfo{pages}{342--352}.
\newblock


\bibitem[Gillet et~al\mbox{.}(2021)]%
        {gillet2021robot}
\bibfield{author}{\bibinfo{person}{Sarah Gillet}, \bibinfo{person}{Ronald Cumbal}, \bibinfo{person}{Andr{\'e} Pereira}, \bibinfo{person}{Jos{\'e} Lopes}, \bibinfo{person}{Olov Engwall}, {and} \bibinfo{person}{Iolanda Leite}.} \bibinfo{year}{2021}\natexlab{}.
\newblock \showarticletitle{Robot gaze can mediate participation imbalance in groups with different skill levels}. In \bibinfo{booktitle}{\emph{Proceedings of the 2021 ACM/IEEE International Conference on Human-Robot Interaction}}. \bibinfo{pages}{303--311}.
\newblock


\bibitem[Jeong et~al\mbox{.}(2023)]%
        {jeong2023robotic}
\bibfield{author}{\bibinfo{person}{Sooyeon Jeong}, \bibinfo{person}{Laura Aymerich-Franch}, \bibinfo{person}{Sharifa Alghowinem}, \bibinfo{person}{Rosalind~W Picard}, \bibinfo{person}{Cynthia~L Breazeal}, {and} \bibinfo{person}{Hae~Won Park}.} \bibinfo{year}{2023}\natexlab{}.
\newblock \showarticletitle{A robotic companion for psychological well-being: A long-term investigation of companionship and therapeutic alliance}. In \bibinfo{booktitle}{\emph{Proceedings of the 2023 ACM/IEEE International Conference on Human-Robot Interaction}}. \bibinfo{pages}{485--494}.
\newblock


\bibitem[Kahn~Jr et~al\mbox{.}(2012)]%
        {kahn2012robovie}
\bibfield{author}{\bibinfo{person}{Peter~H Kahn~Jr}, \bibinfo{person}{Takayuki Kanda}, \bibinfo{person}{Hiroshi Ishiguro}, \bibinfo{person}{Nathan~G Freier}, \bibinfo{person}{Rachel~L Severson}, \bibinfo{person}{Brian~T Gill}, \bibinfo{person}{Jolina~H Ruckert}, {and} \bibinfo{person}{Solace Shen}.} \bibinfo{year}{2012}\natexlab{}.
\newblock \showarticletitle{“Robovie, you'll have to go into the closet now”: Children's social and moral relationships with a humanoid robot.}
\newblock \bibinfo{journal}{\emph{Developmental psychology}} \bibinfo{volume}{48}, \bibinfo{number}{2} (\bibinfo{year}{2012}), \bibinfo{pages}{303}.
\newblock


\bibitem[Komatsu et~al\mbox{.}(2021)]%
        {komatsu2021blaming}
\bibfield{author}{\bibinfo{person}{Takanori Komatsu}, \bibinfo{person}{Bertram~F Malle}, {and} \bibinfo{person}{Matthias Scheutz}.} \bibinfo{year}{2021}\natexlab{}.
\newblock \showarticletitle{Blaming the reluctant robot: parallel blame judgments for robots in moral dilemmas across US and Japan}. In \bibinfo{booktitle}{\emph{Proceedings of the 2021 ACM/IEEE International Conference on Human-Robot Interaction}}. \bibinfo{pages}{63--72}.
\newblock


\bibitem[Ligthart et~al\mbox{.}(2023)]%
        {ligthart2023design}
\bibfield{author}{\bibinfo{person}{Mike~EU Ligthart}, \bibinfo{person}{Simone~M de Droog}, \bibinfo{person}{Marianne Bossema}, \bibinfo{person}{Lamia Elloumi}, \bibinfo{person}{Kees Hoogland}, \bibinfo{person}{Matthijs~HJ Smakman}, \bibinfo{person}{Koen~V Hindriks}, {and} \bibinfo{person}{Somaya Ben~Allouch}.} \bibinfo{year}{2023}\natexlab{}.
\newblock \showarticletitle{Design Specifications for a Social Robot Math Tutor}. In \bibinfo{booktitle}{\emph{Proceedings of the 2023 ACM/IEEE International Conference on Human-Robot Interaction}}. \bibinfo{pages}{321--330}.
\newblock


\bibitem[Lin et~al\mbox{.}(2022)]%
        {lin2022benefits}
\bibfield{author}{\bibinfo{person}{Ting-Han Lin}, \bibinfo{person}{Spencer Ng}, {and} \bibinfo{person}{Sarah Sebo}.} \bibinfo{year}{2022}\natexlab{}.
\newblock \showarticletitle{Benefits of an Interactive Robot Character in Immersive Puzzle Games}. In \bibinfo{booktitle}{\emph{2022 31st IEEE International Conference on Robot and Human Interactive Communication (RO-MAN)}}. IEEE, \bibinfo{pages}{37--44}.
\newblock


\bibitem[Martelaro et~al\mbox{.}(2016)]%
        {martelaro2016tell}
\bibfield{author}{\bibinfo{person}{Nikolas Martelaro}, \bibinfo{person}{Victoria~C Nneji}, \bibinfo{person}{Wendy Ju}, {and} \bibinfo{person}{Pamela Hinds}.} \bibinfo{year}{2016}\natexlab{}.
\newblock \showarticletitle{Tell me more designing hri to encourage more trust, disclosure, and companionship}. In \bibinfo{booktitle}{\emph{2016 11th ACM/IEEE International Conference on Human-Robot Interaction (HRI)}}. IEEE, \bibinfo{pages}{181--188}.
\newblock


\bibitem[Mavrogiannis et~al\mbox{.}(2022)]%
        {mavrogiannis2022winding}
\bibfield{author}{\bibinfo{person}{Christoforos Mavrogiannis}, \bibinfo{person}{Krishna Balasubramanian}, \bibinfo{person}{Sriyash Poddar}, \bibinfo{person}{Anush Gandra}, {and} \bibinfo{person}{Siddhartha~S Srinivasa}.} \bibinfo{year}{2022}\natexlab{}.
\newblock \showarticletitle{Winding Through: Crowd Navigation via Topological Invariance}.
\newblock \bibinfo{journal}{\emph{IEEE Robotics and Automation Letters}} \bibinfo{volume}{8}, \bibinfo{number}{1} (\bibinfo{year}{2022}), \bibinfo{pages}{121--128}.
\newblock


\bibitem[Mazursky et~al\mbox{.}(2022)]%
        {mazursky2022physical}
\bibfield{author}{\bibinfo{person}{Alex Mazursky}, \bibinfo{person}{Madeleine DeVoe}, {and} \bibinfo{person}{Sarah Sebo}.} \bibinfo{year}{2022}\natexlab{}.
\newblock \showarticletitle{Physical Touch from a Robot Caregiver: Examining Factors that Shape Patient Experience}. In \bibinfo{booktitle}{\emph{2022 31st IEEE International Conference on Robot and Human Interactive Communication (RO-MAN)}}. IEEE, \bibinfo{pages}{1578--1585}.
\newblock


\bibitem[Naggita et~al\mbox{.}(2022)]%
        {naggita2022parental}
\bibfield{author}{\bibinfo{person}{Keziah Naggita}, \bibinfo{person}{Elsa Athiley}, \bibinfo{person}{Beza Desta}, {and} \bibinfo{person}{Sarah Sebo}.} \bibinfo{year}{2022}\natexlab{}.
\newblock \showarticletitle{Parental Responses to Aggressive Child Behavior towards Robots, Smart Speakers, and Tablets}. In \bibinfo{booktitle}{\emph{2022 31st IEEE International Conference on Robot and Human Interactive Communication (RO-MAN)}}. IEEE, \bibinfo{pages}{337--344}.
\newblock


\bibitem[Nanavati et~al\mbox{.}(2023)]%
        {nanavati2023design}
\bibfield{author}{\bibinfo{person}{Amal Nanavati}, \bibinfo{person}{Patricia Alves-Oliveira}, \bibinfo{person}{Tyler Schrenk}, \bibinfo{person}{Ethan~K Gordon}, \bibinfo{person}{Maya Cakmak}, {and} \bibinfo{person}{Siddhartha~S Srinivasa}.} \bibinfo{year}{2023}\natexlab{}.
\newblock \showarticletitle{Design principles for robot-assisted feeding in social contexts}. In \bibinfo{booktitle}{\emph{Proceedings of the 2023 ACM/IEEE International Conference on Human-Robot Interaction}}. \bibinfo{pages}{24--33}.
\newblock


\bibitem[Natarajan et~al\mbox{.}(2023)]%
        {natarajan2023human}
\bibfield{author}{\bibinfo{person}{Manisha Natarajan}, \bibinfo{person}{Esmaeil Seraj}, \bibinfo{person}{Batuhan Altundas}, \bibinfo{person}{Rohan Paleja}, \bibinfo{person}{Sean Ye}, \bibinfo{person}{Letian Chen}, \bibinfo{person}{Reed Jensen}, \bibinfo{person}{Kimberlee~Chestnut Chang}, {and} \bibinfo{person}{Matthew Gombolay}.} \bibinfo{year}{2023}\natexlab{}.
\newblock \showarticletitle{Human-robot teaming: grand challenges}.
\newblock \bibinfo{journal}{\emph{Current Robotics Reports}} \bibinfo{volume}{4}, \bibinfo{number}{3} (\bibinfo{year}{2023}), \bibinfo{pages}{81--100}.
\newblock


\bibitem[Pelikan et~al\mbox{.}(2020)]%
        {pelikan2020you}
\bibfield{author}{\bibinfo{person}{Hannah~RM Pelikan}, \bibinfo{person}{Mathias Broth}, {and} \bibinfo{person}{Leelo Keevallik}.} \bibinfo{year}{2020}\natexlab{}.
\newblock \showarticletitle{" Are you sad, Cozmo?" How humans make sense of a home robot's emotion displays}. In \bibinfo{booktitle}{\emph{Proceedings of the 2020 ACM/IEEE International Conference on Human-Robot Interaction}}. \bibinfo{pages}{461--470}.
\newblock


\bibitem[Reig et~al\mbox{.}(2021)]%
        {reig2021flailing}
\bibfield{author}{\bibinfo{person}{Samantha Reig}, \bibinfo{person}{Elizabeth~J Carter}, \bibinfo{person}{Terrence Fong}, \bibinfo{person}{Jodi Forlizzi}, {and} \bibinfo{person}{Aaron Steinfeld}.} \bibinfo{year}{2021}\natexlab{}.
\newblock \showarticletitle{Flailing, hailing, prevailing: Perceptions of multi-robot failure recovery strategies}. In \bibinfo{booktitle}{\emph{Proceedings of the 2021 ACM/IEEE International Conference on Human-Robot Interaction}}. \bibinfo{pages}{158--167}.
\newblock


\bibitem[Rifinski et~al\mbox{.}(2021)]%
        {rifinski2021human}
\bibfield{author}{\bibinfo{person}{Danielle Rifinski}, \bibinfo{person}{Hadas Erel}, \bibinfo{person}{Adi Feiner}, \bibinfo{person}{Guy Hoffman}, {and} \bibinfo{person}{Oren Zuckerman}.} \bibinfo{year}{2021}\natexlab{}.
\newblock \showarticletitle{Human-human-robot interaction: robotic object’s responsive gestures improve interpersonal evaluation in human interaction}.
\newblock \bibinfo{journal}{\emph{Human--Computer Interaction}} \bibinfo{volume}{36}, \bibinfo{number}{4} (\bibinfo{year}{2021}), \bibinfo{pages}{333--359}.
\newblock


\bibitem[Robinette et~al\mbox{.}(2016)]%
        {robinette2016overtrust}
\bibfield{author}{\bibinfo{person}{Paul Robinette}, \bibinfo{person}{Wenchen Li}, \bibinfo{person}{Robert Allen}, \bibinfo{person}{Ayanna~M Howard}, {and} \bibinfo{person}{Alan~R Wagner}.} \bibinfo{year}{2016}\natexlab{}.
\newblock \showarticletitle{Overtrust of robots in emergency evacuation scenarios}. In \bibinfo{booktitle}{\emph{2016 11th ACM/IEEE International Conference on Human-Robot Interaction (HRI)}}. IEEE, \bibinfo{pages}{101--108}.
\newblock


\bibitem[Tennent et~al\mbox{.}(2019)]%
        {tennent2019micbot}
\bibfield{author}{\bibinfo{person}{Hamish Tennent}, \bibinfo{person}{Solace Shen}, {and} \bibinfo{person}{Malte Jung}.} \bibinfo{year}{2019}\natexlab{}.
\newblock \showarticletitle{Micbot: A peripheral robotic object to shape conversational dynamics and team performance}. In \bibinfo{booktitle}{\emph{2019 14th ACM/IEEE International Conference on Human-Robot Interaction (HRI)}}. IEEE, \bibinfo{pages}{133--142}.
\newblock


\bibitem[Winkle et~al\mbox{.}(2023)]%
        {winkle2023feminist}
\bibfield{author}{\bibinfo{person}{Katie Winkle}, \bibinfo{person}{Donald McMillan}, \bibinfo{person}{Maria Arnelid}, \bibinfo{person}{Katherine Harrison}, \bibinfo{person}{Madeline Balaam}, \bibinfo{person}{Ericka Johnson}, {and} \bibinfo{person}{Iolanda Leite}.} \bibinfo{year}{2023}\natexlab{}.
\newblock \showarticletitle{Feminist human-robot interaction: Disentangling power, principles and practice for better, more ethical HRI}. In \bibinfo{booktitle}{\emph{Proceedings of the 2023 ACM/IEEE International Conference on Human-Robot Interaction}}. \bibinfo{pages}{72--82}.
\newblock


\end{thebibliography}


\end{document}